\documentclass[twocolumn]{IEEEtran}
\usepackage{graphicx}
\usepackage{subfigure}
\usepackage{hyperref}
\usepackage{algorithm,calc}
\usepackage[noend]{algpseudocode}
\usepackage{flushend}

\makeatletter
\def\BState{\State\hskip-\ALG@thistlm}
\makeatother

\newcommand{\mathbox}[3][r]{\makebox[\widthof{$#2$}][#1]{$#3$}}

\usepackage{authblk}
\usepackage{epstopdf}

\usepackage{amsmath}  % needed for \tfrac, \bmatrix, etc.
\usepackage{amsfonts} % needed for bold Greek, Fraktur, and blackboard bold
\usepackage{graphicx} % needed for figures

\begin{document}

\pdfoutput=1

\title{Stepwise regression for unsupervised  learning}
\author{
Jonathan Landy\thanks{The author is a member of the data-science group at Square Inc., which is located in San Francisco, CA.  (email:  jslandy@squareup.com)}
}

\maketitle % title page is now complete

\begin{abstract}
I consider unsupervised extensions of the fast stepwise linear regression algorithm \cite{efroymson1960multiple}.  These extensions allow one to efficiently identify highly-representative feature variable subsets within a given set of jointly distributed variables.  This in turn allows for the efficient dimensional reduction of large data sets via the removal of redundant features.  Fast search is effected here through the avoidance of repeat computations across trial fits, allowing for a full representative-importance ranking of a set of feature variables to be carried out in $O(n^2 m)$ time, where $n$ is the number of variables and $m$ is the number of data samples available.  This runtime complexity matches that needed to carry out a single regression and is $O(n^2)$ faster than that of naive implementations.  I present pseudocode suitable for efficient forward, reverse, and forward-reverse unsupervised feature selection.  To illustrate the algorithm's application, I apply it to the problem of identifying representative stocks within a given financial market index -- a challenge relevant to the design of Exchange Traded Funds (ETFs).  I also characterize the growth of numerical error with iteration step in these algorithms, and finally demonstrate and rationalize the observation that the forward and reverse algorithms return exactly inverted feature orderings in the weakly-correlated feature set regime.
\end{abstract}

\section{Introduction}

Raw data sets sometimes contain a significant number of redundant or nearly-redundant measurement features -- feature variables that can  be accurately predicted given the values taken on by each of the others. The pruning of redundant features from a large data set can result in a number of desirable consequences.  Most obviously, their removal allows for a smaller, compressed representation of the original data set.  Redundant feature removal can also result in improved interpretability of data sets as well as reduced training time, size, and over-fitting of models built on top of them \cite{bruckstein2009sparse, guyon2003introduction}.  These benefits motivate the search for general, automated feature selection algorithms that can be applied at scale.

The fundamental challenge associated with selecting optimal feature subsets is that the number of candidates of a given size can be very large -- given a set of $n$ feature vectors, there are $n \choose k$ vector subsets of size $k$.   This combinatorial scaling implies that we must often trade-off guaranteed optimality for improved run time in our search strategies.  In the supervised regime -- where one attempts to obtain a best fit to a target vector $y$ using at most $k$ of the available predictors - two linear regression-based approaches have proven to provide an excellent balance of speed and accuracy:  First,  a greedy, step-wise regression method has been introduced that can be made to run very quickly via certain matrix inverse update identities \cite{efroymson1960multiple}.  In the forward variant of this algorithm, features are added to the predictor set one at a time.  At each step, the feature added is that which results in the best improvement to the regression on the target variable.  In this way, an approximate importance ranking of the full set of features is  obtained -- with the first feature added considered most important, etc.  Dropping all but the top $k$ features provides an approximate solution to the optimal selection problem.  The second method that has been pursued is an approximate mapping onto a convex optimization problem \cite{bruckstein2009sparse}.  This approach does not provide a ranking of the features, but instead simply outputs the feature subset that minimizes the sum of the regression squared error and a sparsity-enforcing regularization term.  Though both algorithms can be carried out in polynomial time,  each is known to provide globally optimal solutions under certain conditions -- a boon given that optimal subset selection is an NP-hard task, in general \cite{natarajan1995sparse}.

In this paper, I consider the challenge of feature selection in the unsupervised regime -- that where one wishes to identify a representative subset of features in a given data set, not having any particular target variable in mind.  A natural strategy for addressing this challenge is to attempt a translation of the successful linear methods noted above to the unsupervised task of feature set partitioning.   Here, the goal is to segment a given feature set into a retained set and a rejected set, with the optimal target rejection set of a given size being that having the least squared error when projected onto its complement -- i.e., that most accurately approximated by a linear fit constructed from the features in the corresponding retained set.  Extensions of the convex mapping approach that address this problem have been detailed elsewhere \cite{masaeli2010convex}.  However,  prior consideration of the unsupervised stepwise approach appears to be limited to a forward selection variant \cite{farahat2013efficient}.  Here, I show that the original supervised formalism \cite{efroymson1960multiple} can be directly extended to the unsupervised regime, and I use this approach to write down and characterize simple implementations of the unsupervised forward, reverse, and hybrid forward-reverse algorithms.  Each provides an efficient routine, specialized for optimization in a different limit:  The forward algorithm provides near optimal solutions when the subset selected is small, while the reverse algorithm performs best in the opposite limit.  Finally, the hybrid forward-reverse algorithm can provide significantly better results than the other two when a subset of intermediate size is selected.  I illustrate the application of the algorithms here and also present pseudo-code implementations for each.

The remainder of the paper proceeds as follows:  In Sec.\ \ref{sec:overview}, I (i) formally define the selection problem considered here and outline the greedy strategies for addressing it, (ii) discuss a naive implementation of the greedy strategies, and (iii) provide an overview of their efficient implementations.  In Sec.\ \ref{sec:algorithm}, I present pseudocode for the forward, reverse, and forward-reverse algorithms.  In Sec.\ \ref{sec:stocks}, I illustrate application of the algorithms through consideration of a stock market fluctuation analysis.   Sec.\ \ref{sec:discussion} contains a brief discussion of the results.  I also include three appendices.  Appendix \ref{sec:derivation} contains derivations relating to the update procedure.  Appendix \ref{sec:error} contains an analysis of the growth in numerical round off error in the greedy algorithms.  Finally, Appendix \ref{sec:relationship} relates to the relationship between the forward and reverse algorithms -- here, I demonstrate that they return exactly opposite orderings in certain limits.

\section{Problem definition and algorithm overview}
\subsection*{Problem statement and greedy solution strategies}
\label{sec:overview}
The unsupervised feature selection problem that I consider here centers around the characterization of a given $n \times m$ data matrix $X$, with element $X_{jk}$ holding the value of the $j$-th feature in the $k$-th  data instance.  For simplicity, I assume throughout this paper that the rows of $X$ are linearly independent, which requires $m > n$, and I also assume that the features have been shifted and normalized to each have mean $0$ and variance $1$.  The features can be formally considered to be jointly distributed random variables.  To score the representative quality of a given feature subset $s$, I take the squared residual error associated with projecting $X$ onto the rows of $s$ -- i.e., the squared error of the best linear fit to the whole data set using only the features within $s$.  This is,
\begin{eqnarray}\label{projection_error}
\mathcal{F}(s) \equiv \left \| X^T - \mathcal{P}^{(s)} X^T  \right \|^2,
\end{eqnarray}
where $\mathcal{P}^{(s)} $ is the projection operator \cite{apostol1969calculus},
\begin{eqnarray}\label{projection_matrix}
\mathcal{P}^{(s)}  = \tilde{X}^T \left (\tilde{X} \tilde{X}^T \right )^{-1} \tilde{X}.
\end{eqnarray}
Here, $\tilde{X}$ is defined analogously to $X$, but with the rows outside of $s$ removed.  Notice that the inverse above exists, given the assumption of linear independence, and that the common normalization of the features results in each being weighted equally in the projection error.

Although linear fits are relatively easy to carry out,  global minimization of (\ref{projection_error}) is challenging due to the combinatorial number of candidate subsets. A greedy search procedure can be applied to obtain low-cost subsets in an efficient manner.  This  approach centers around iterative addition or removal of features from $s$, one at a time, at each step selecting that single feature that results in the best outcome.  In the forward process, one begins with no features in $s$.  Features are then iteratively added in, one at a time, always selecting that feature that reduces $\mathcal{F}$ the most.  That is, at a given step one identifies and adds to $s$ that feature $d$ satisfying
\begin{eqnarray} \label{greedy_problem}
d = \text{argmin}_j \ \mathcal{F}(s + \{j\}).
\end{eqnarray}
 In the reverse process, one begins with all features in $s$.  Features are then iteratively removed from $s$, one at a time, at each step selecting and removing that feature $d$ satisfying
\begin{eqnarray} \label{greedy_problem_rev}
d = \text{argmin}_j \ \mathcal{F}(s - \{j\}).
\end{eqnarray}
The forward-reverse greedy algorithm is a hybrid of the two above:  One sometimes adds features to $s$ and sometimes removes them.   Various protocols can be considered for determining when a step forward is taken, rather than a step backward -- I discuss and outline one choice below.  

The change in $\mathcal{F}$ that occurs in a given step of these algorithms can be decomposed into two parts:  The change in the projection error of $d$ and the change in the projection error associated with the remaining features  outside of $s$ -- when $d$ is either added or removed from $s$, the fit to these other features may improve or worsen.  Both contributions need to be considered in order to select the optimal feature at each step.
 
The forward and reverse greedy algorithms are very similar to gradient descent, in that one always takes a step in the direction of least cost.  The subsets obtained in this way provide approximate minimizations of  (\ref{projection_error}).   These can be obtained relatively quickly because the greedy search protocol is restricted.  For example, in the forward approach, once a feature $d$ is added to $s$, the algorithm will only explore predictor subsets that include it from that point forward.   Although restricted, the  greedy strategies contain mechanisms for identifying competitive subsets: If clusters of the features exist,  with each feature in a given cluster very similar, the algorithms will generally work towards identifying an $s$ that contains a single representative from each.  For example, in the reverse process, pruning features from a cluster of very similar features will come at very little cost, since each can be well-predicted from each of the others.  However, a significant cost would be incurred upon removal of the last feature of a cluster from $s$, as doing so would simultaneously raise the loss on the full similar subset. Because of this, one can expect the greedy pruning process to effectively work towards selecting single representative features from each well-defined cluster that appears in the predictor set -- an approach that intuitively suggests we can expect well-optimized feature orderings.  Note that the forward-reverse hybrid approach allows for a more exhaustive search, and for this reason often identifies lower cost subsets of a given size.  However, these improvements come at the cost of increased run time.   

\subsection*{Naive greedy implementations} 
To initiate the reverse greedy process described above, the first step one must take is to determine which of the $n$ features should be removed from $s$ first.  This is that feature that is most easily predicted from each of the others.   To identify this feature, a natural first approach would be to develop individual linear fits to each of the $n$ separate features, in each case using the remaining $(n-1)$ features as predictor variables.  For example, one would first fit the feature $x_1$ using the variables $\{x_2,  x_3,\ldots, x_n\}$.  Next, one would fit $x_2$, using the predictors $\{x_1, x_3, \ldots, x_n\}$, etc.  The feature whose regression returned the smallest projection error would then be that most accurately predicted by each of the other features.  With this feature identified, it can then be pruned from $s$.  In the next iteration of the process, one additional feature must be removed from $s$.  This will be that satisfying (\ref{greedy_problem_rev}), leaving the projection error on the full pruned set -- now of size two -- minimized.  The process would then continue from there.

The problem with this naive implementation strategy is that it will run very slowly at large $n$:   Evaluation of $\mathcal{F}$ for a single subset requires evaluation of the projection operator (\ref{projection_matrix}), which requires $O(n^2m)$ computations.  Each iteration of the algorithm requires $O(n)$ candidate subsets to be considered, which will therefore require $O(n) \times O(n^2 m) = O(n^3 m)$ time.  The full runtime complexity required to obtain a ranking of the full set of $n$ features then scales like $O(n^4 m)$.  This scaling precludes applications of the naive approach at large $n$.  
 
\subsection*{Efficient greedy implementations}  
 
The fast, stepwise linear regression algorithm \cite{efroymson1960multiple} is typically applied to the problem of minimizing the squared regression error of a fixed target variable.  However, I show here that the algorithm can be simply extended to also allow for the efficient implementation of the greedy minimization of (\ref{projection_error}).  The crux of this algorithm is an application of the Woodbury matrix inverse update formula \cite{woodbury1950inverting}.  This identity allows one to efficiently update the matrix inverse that appears in (\ref{projection_error}) as features are either added or subtracted from $s$ -- thereby avoiding a costly, fresh inverse reevaluation with each iteration.  The result is a significant speedup.

To simplify the presentation of the procedure, it is convenient to introduce a few matrices related to the feature correlation matrix, $M$, which has $i-j$ component
\begin{eqnarray} \label{correlation}
M_{ij} \equiv X_i \cdot X_j^T.
\end{eqnarray}
Recalling the assumed normalization of the features, and writing $\mathcal{F}_j$ for the squared projection error of the $j$-th feature, combining  (\ref{projection_error}),  (\ref{projection_matrix}), and  (\ref{correlation}) gives
\begin{eqnarray}\label{rjsquared}
\mathcal{F}_j = 1 -  M_j \cdot R \cdot M_j^T.
\end{eqnarray}
Here, $M_j$ is the $j$-th row of the original correlation matrix and $R$ is the matrix having elements
\begin{eqnarray} \label{minvdef}
R_{ij} = 
\begin{cases}
0, & \text{if $i$ or $j \not \in s$} \\
 \left (\tilde{X} \tilde{X}^T \right )^{-1}_{ij}, & \text{else}.
\end{cases}
\end{eqnarray}
Notice that projecting $R$ onto $s$ gives the inverse of $\tilde{M}$, the projection of $M$ onto $s$.   Working with $R$ rather than $\tilde{M}^{-1}$ allows one to update this matrix in place --  i.e., it avoids the need to copy the inverse over to one having a different dimension with each iteration.   The two final matrices that must be introduced are $U$ and $D$.  These have elements
\begin{eqnarray} \label{Udef}
U_{ij} \equiv M_i \cdot R \cdot M_j^T,
\end{eqnarray}
and
\begin{eqnarray}
\label{Ddef}
D_{ij} \equiv R_i \cdot M_j.
\end{eqnarray}

I outline a derivation for the Woodbury update expressions for (\ref{minvdef}) in appendix \ref{sec:derivation}.   These read as follows:  When feature $d$ is removed from $s$, $R$ is updated to 
\begin{eqnarray}\label{updateformula}
R  \leftarrow  R  - \frac{R_d \otimes R_d }{R_{dd}},
\end{eqnarray}
where $\otimes$ represents the outer product and $R_d$ is the $d$-th row of $R$.  Similarly, when feature $d$ is added to $s$, $R$ is updated to
\begin{eqnarray} \label{forward_inverse_update}
R \leftarrow R + \frac{\left (D^T_d- e_d \right) \otimes \left (D^T_d - e_d \right) }{1 - U_{dd}},
\end{eqnarray}
where $e_d$ is the vector that is $1$ at index $d$ and $0$ elsewhere.   The update formulas for $U$ and $D$ are obtained by combining their definitions with the update formulas for $R$:   If feature $d$ is removed from $s$, combining  (\ref{Udef}), (\ref{Ddef}), and (\ref{updateformula})  gives
\begin{eqnarray}
\label{reverse_u_update}
U &\leftarrow & U - \frac{D_d \otimes D_d}{R_{dd}},  \\
\label{reverse_d_update}
D &\leftarrow & D -  \frac{R_d \otimes D_d}{R_{dd}}.
\end{eqnarray}
Similarly, when feature $d$ is added to $s$, combining (\ref{Udef}),  (\ref{Ddef}), and  (\ref{forward_inverse_update}) gives
\begin{eqnarray}
\label{forward_u_update}
U &\leftarrow& U + \frac{(U_d - M_d) \otimes (U_d - M_d)}{1 - U_{dd}} \\
\label{forward_d_update}
D &\leftarrow& D + \frac{ (D_d^T - e_d) \otimes (U_d - M_d)}{1 - U_{dd}}.
\end{eqnarray}

The matrices $M$, $R$, $U$, and $D$ above allow for a speedup in the greedy selection process because they contain all of the information needed  to determine the change in projection error associated with single variable adjustments to $s$.   For example, if feature $d$ is removed from $s$, combining  (\ref{rjsquared}) and (\ref{updateformula}) gives
\begin{eqnarray} \nonumber \label{projection_update} 
\mathcal{F}_j &\leftarrow & 1 -  M_j \cdot \left (  R - \frac{R_d \otimes R_d }{R_{dd}} \right) \cdot M_j^T \\ \nonumber
&=&  \mathcal{F}_j + \frac{ \left ( R_d  \cdot M_j  \right )^2}{R_{dd}} \\
&\equiv & \mathcal{F}_j  +  \frac{ D_{dj} ^2}{R_{dd}}.
\end{eqnarray}
Summing over all $j$ gives the full projection error update that results from the removal of feature $d$,
\begin{eqnarray}\label{cost_update}
\mathcal{F} \leftarrow \mathcal{F} +  \frac{\left \| D_d \right \|^2}{R_{dd}}.
\end{eqnarray}
If $R$ and $D$ are evaluated and updated as the algorithm is carried out, the right side of this equation can be evaluated to check how $\mathcal{F}$ would change with the removal of $d$ -- all without actually evaluating the new projection operator. Further, this check simply involves the evaluation of a dot product and can be carried out in $O(n)$ time -- a significant speedup over the $O(n^2 m)$ time taken in the naive approach to score a given candidate feature.  Scoring all features then requires only $O(n^2)$ time per iteration.  Once the optimal feature is identified, it can be removed from $s$, and the updates (\ref{updateformula}), (\ref{reverse_u_update}) and (\ref{reverse_d_update}) can be applied.  Each of the three updates also requires only $O(n^2)$ computations.  Once these have been carried out, the process can be iterated from there.  

Similarly, if in the forward process the feature $d$ is added to $s$, combining (\ref{rjsquared}) and (\ref{forward_inverse_update}) gives
\begin{eqnarray} \nonumber \label{projection_update_forward}
\mathcal{F}_j &\leftarrow & 1  - M_j \cdot \left ( R + \frac{\left (D^T_d- e_d \right) \otimes \left (D^T_d - e_d \right) }{1 - U_{dd}} \right) \cdot M_j^T\\ 
& \equiv & \mathcal{F}_j  - \frac{\left ( U_{jd} - M_{jd} \right )^2 }{1 - U_{dd}}. 
\end{eqnarray}
Summing over all $j$ gives the full cost function increase,
\begin{eqnarray}\label{forward_error_update}
\mathcal{F} \leftarrow \mathcal{F} - \frac{\left \| U_d - M_d \right \|^2}{1 - U_{dd}}.
\end{eqnarray}
Using this expression, the optimal feature for addition can again be evaluated in $O(n^2)$ time.  Once identified, the updates (\ref{forward_inverse_update}), (\ref{forward_u_update}), and (\ref{forward_d_update}) can be carried out, and the process continued.  Again, a full iteration only requires $O(n^2)$ computations.

In summary then, the processes above each allow for either a forward or a reverse step to be carried out in $O(n^2)$ time.  A full ranking of the $n$ features can then be carried out in $O(n^3)$ time.   Because $m > n$,  this is necessarily smaller than initial the time required to evaluate $M$, $O(n^2 m)$ time.  This initial computation sets the scaling for a full run of the fast forward and reverse greedy algorithms.   The result is a process that is a factor of $O(n^2)$ faster than the naive implementations discussed above.  For example, if $n \sim O(10^3)$, this means that the inverse update procedure provides an $O(10^6)$ speedup,  which is quite significant.  

The matrices $R$, $U$, and $D$ --  and their update formulas presented  above -- are identical to those evaluated in the supervised stepwise linear regression algorithm \cite{efroymson1960multiple}.  The central difference between the supervised algorithm and those considered here is the cost function that determines the optimal feature for selection at each step.  In the supervised case, the regression error on the target variable defines the cost, whereas in the unsupervised case, the cost is given by the squared regression error on all features not in $s$ -- (\ref{cost_update}) in the reverse case and (\ref{forward_error_update}) in the forward case.  The results above show that this extension can be carried out without increasing the runtime complexity of the algorithms.

The next section contains summary pseudocode for each algorithm.

\section{Algorithm pseudocode}
\label{sec:algorithm}

\subsection{Forward process}

The forward selection algorithm repeatedly applies (\ref{forward_error_update}) to evaluate the cost of selecting any given feature for inclusion in $s$ at a given step.  Once the optimal feature for an iteration is identified, it is added to $s$, the current squared projection error (\ref{projection_error}) is evaluated and stored, and the matrices $R$ and $U$ are updated -- again, using the forward update expressions (\ref{forward_inverse_update}) and (\ref{forward_u_update}).  The summary pseudocode for this process is given in Algorithm \ref{alg:fsfs}.  As written, this accepts a data matrix $X$ and returns a list of the features, ordered as selected by the algorithm.  The algorithm also returns a list of the squared projection error realized at each iteration of the algorithm.

\begin{algorithm} [t]
\caption{Unsupervised Forward Selection (UFS).  The components are 1) $X$, the input data matrix, 2-4) $M$, $R$, and $U$ --  matrices defined in (\ref{correlation}), (\ref{minvdef}), and (\ref{Udef}),  5) $added\_list$, a list of the features, ordered from most important to least, and 6) $f\_list$, a list whose $i$-th element contains the value of (\ref{projection_error}) that results when only the first $i$ features of $added\_list$ are retained. }\label{alg:fsfs}
\begin{algorithmic}[1]
\Procedure{UFS}{$X$}
\State $M = corr(X)$
\State $R = 0_{len(M) \times len(M)}$
\State $U = 0_{len(M) \times len(M)}$
\State $added\_list, f\_list = [], [len(M)]$
\For{$i = 1, 2, \ldots, len(M)$}
\State $d , v = \text{argmax}_{j \not \in added\_list}   \frac{1}{1-U_{jj}} \| U_j - M_j\|^2$
\State $R \mathrel{+}=  \frac{1}{1 - U_{dd}} ( M^{-1} \cdot M_d - e_d ) \otimes ( M^{-1} \cdot M_d - e_d ) $
\State $ \mathbox{R}{U} \mathrel{+}=  \frac{1}{1 - U_{dd}} (U_d - M_d) \otimes (U_d - M_d) $
\State $added\_list.append(d)$
\State $f\_list.append(r2\_list[-1] - v)$
\EndFor
\State \textbf{return} $added\_list, f\_list[1:]$
\EndProcedure
\end{algorithmic}
\end{algorithm}

\subsection{Reverse process}

This reverse procedure is very similar to the forward selection process, but begins with all features included.  The algorithm then iteratively removes features from $s$ one at a time.  The optimal feature for removal is identified using (\ref{cost_update}).  Once this is found, the selected feature set, cost function, and the matrices $R$ and $D$ are updated -- the latter using (\ref{updateformula}) and (\ref{reverse_d_update}).   The summary pseudo-code for the reverse process is given in Algorithm \ref{alg:urs}.   As written, the algorithm returns a list of the features in the order in which they were pruned from $s$, as well as a list of the cost function that results at each step of the algorithm.

\begin{algorithm} [t]
\caption{Unsupervised Reverse Selection (URS).  The components are 1) $X$, the input data matrix, 2-4) the matrices $M$, $R$, and $D$ -- defined in (\ref{correlation}), (\ref{minvdef}), and (\ref{Ddef}), 5) $pruned\_list$, a list of the features, ordered from least important to most, and  6) $f\_list$, a list whose $i$-th element contains the value of (\ref{projection_error}) that results when only the first $i$ features of $pruned\_list$ are removed from the feature set. }\label{alg:urs}
\begin{algorithmic}[1]
\Procedure{URS}{$X$}
\State $M = corr(X)$
\State $R = inv(M)$
\State $D = I(len(M))$
\State $pruned\_list, f\_list= [], [0]$
\For{$i = 1, 2, \ldots, len(M)$}
\State $d, v = \text{argmin}_{j \not \in pruned\_list}   \frac{1}{R_{jj}} \| D_j \|^2$
\State $ \mathbox{R}{D} \mathrel{-}=  \frac{1}{R_{dd}} (R_d \otimes D_d)$
\State $R \mathrel{-}=  \frac{1}{R_{dd}}(R_d \otimes  R_d) $
\State $pruned\_list.append(d)$
\State $f\_list.append(f\_list[-1] + v)$
\EndFor
\State \textbf{return} $pruned\_list, f\_list[1:]$
\EndProcedure
\end{algorithmic}
\end{algorithm}

\subsection{Forward-Reverse selection}
Many variants of the forward-reverse selection process can be considered.  In general, each presents a strategy that determines when a forward step is taken and when a reverse step is taken in the greedy process.  Allowing for removal of features from the retained set $s$ is useful, as a feature that can be very useful when very few features are included may represent a poor feature for inclusion once a number of others have been added to $s$.  

One variant of the hybrid algorithm in presented in Algorithm \ref{alg:frsfs}.  As written, this approach always proceeds by taking $steps$ forward steps followed by $steps -1$ backward steps.  In this way, the algorithm proceeds to move one effective step forward with each iteration of the algorithm, but does so in a seesaw fashion -- allowing for features to be included in the forward process but then removed if it turns out they are suboptimal once others are included.  In practice, I have found setting $steps >2$ tends to produce smoother results than those obtained at $steps = 2$, in the sense that the best cost (\ref{projection_error}) identified tends to vary more smoothly with size $\vert s \vert$.  Increasing the $steps$ parameter requires increased computational time.

\begin{algorithm} [h!]
\caption{Unsupervised Forward Reverse Selection (UFRS).  The components are 1) $X$, the input data matrix, 2-5) the matrices $M$, $R$, $U$ and $D$ -- defined in (\ref{correlation}), (\ref{minvdef}),  (\ref{Udef})  and (\ref{Ddef}), and 6) $opt\_sets$, a dictionary that stores the optimal subset and cost identified for each subset size $\vert s \vert$.  The parameter $steps$ sets how many forward steps are taken with each iteration. }\label{alg:frsfs}
%\begin{spacing}{1.1}
\begin{algorithmic}[1]
\Function{UFRS}{$X$, $steps=2$}

\Function{$update\_opts$}{$added\_indices$, $f$}
\State $k = len(added\_indices)$
\If {$opt\_sets[k][``f"] > f$}
 \State $opt\_sets[k][``set"] = added\_indices.copy()$
  \State $opt\_sets[k][``f"] = f$
\EndIf
\EndFunction

\Function{$forward\_step()$}{}
\If {$len(added\_indices) == n$}
\State \textbf{return}
\EndIf
\State $d, v= \text{argmax}_{j \not \in added\_indices}   \frac{1}{1-U_{jj}} \| U_j - M_j\|^2$
\State $R \mathrel{+}=  \frac{1}{1 - U_{dd}} (D_d^T -  e_d ) \otimes ( D_d^T - e_d ) $
\State $ \mathbox{R}{D} \mathrel{+}=  \frac{1}{1 - U_{dd}} (D_d^T - I_d) \otimes (U_d - M_d) $
\State $ \mathbox{R}{U} \mathrel{+}=  \frac{1}{1 - U_{dd}} (U_d - M_d) \otimes (U_d - M_d) $
\State$f  \mathrel{-}=  v$
\State$added\_indices.add(d)$
\State $update\_opts(added\_indices, f)$
\EndFunction

\Function{$reverse\_step()$}{}
\State $d, v = \text{argmin}_{j \in added\_indices}   \frac{1}{R_{jj}} \| D_j \|^2$
\State $U \mathrel{-}=  \frac{1}{R_{dd}} (D_d \otimes D_d)$
\State $ \mathbox{R}{D} \mathrel{-}=  \frac{1}{R_{dd}} (R_d \otimes D_d)$
\State $R \mathrel{-}=  \frac{1}{R_{dd}}(R_d \otimes  R_d) $
\State $f  \mathrel{+}=  v$
\State $added\_indices.remove(d)$
\State $update\_opts(added\_indices, f)$

\EndFunction

\State $M = corr(X)$
\State $n = len(M)$
\State $I = I_{n \times n}$
\State $R, U, D = 0_{n \times n}, 0_{n \times n}, 0_{n \times n}$
\State $f = len(M)$
\State $added\_indices, opt\_sets= \{ \}, \{ \}$
\For{$i = 1, 2, \ldots, n$}
\State $opt\_sets[i] = \{``set": \{\}, ``f": n \}$
\EndFor
\For{$start = 0, 1, \ldots, n-1$}
\For{$i = 1, 2, \ldots, min(steps, n-start)$}
\State{$forward\_step()$}
\EndFor
\For{$i = 1, 2, \ldots, min(steps, n-start) - 1$}
\State{$reverse\_step()$}
\EndFor
\EndFor
\State \textbf{return} $opt\_sets$
\EndFunction
\end{algorithmic}
%\end{spacing}
\end{algorithm}

\section{Selecting representative stocks}
\label{sec:stocks}

\begin{figure*}[t]\scalebox{0.6}
{\includegraphics[angle=0]{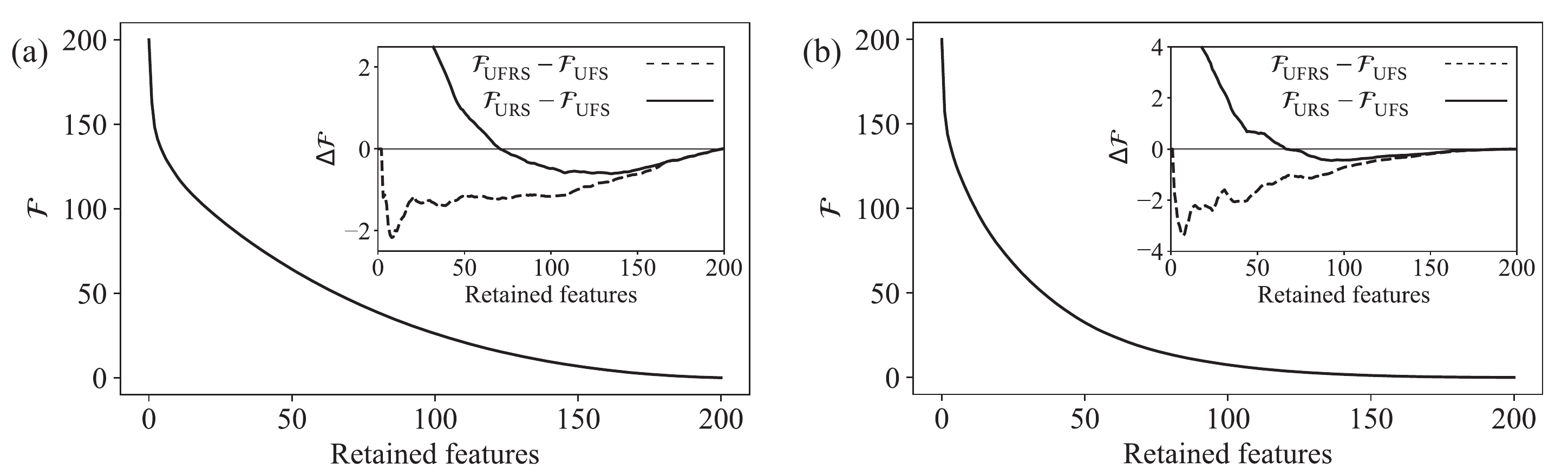}}
\caption{\label{fig:stocks} Results of an unsupervised feature selection analysis of the daily percentage lifts of 200 health-care related stocks traded on NASDAQ throughout 2016.  In main figures are shown the cost function $\mathcal{F}$ associated with retaining only the top $x$ features returned by the unsupervised forward selection algorithm.  Insets show the decrease in cost relative to this associated with switching to the results returned by the UFRS algorithm (dashed, $steps \to 50$) and URS algorithm (solid). (a)  Results of the analysis considering the daily percentage change for each stock over each of the 252 trading days of 2016.  (b)  Results obtained after applying a five-day moving average transform to the raw lift data.  This daily averaging tends to suppress idiosyncratic daily noise.}
\end{figure*} 

Stock Index Exchange Traded Funds (ETFs) are funds that own a portfolio of stocks that together are selected so as to track a complete index, or sector of the market.  These funds divide themselves into shares, allowing for individuals to buy and sell partial ownership of the funds through brokers, much like any other public company stock.  The value of an ETF share is locked to the underlying value of its target index through portfolio transparency, which enables financial institutions to carry out arbitrage conversions between an ETF's shares and the individual stocks comprising its portfolio.  Such conversions will be carried out whenever sufficiently large price discrepancies arise between the two, resulting in price discrepancy suppression.  Unfortunately, these arbitrage trades become more costly to carry out as the number of stocks in an ETF grows, resulting in a breakdown of the price-locking mechanism.  To mitigate this issue, ETFs targeting large indices are now often developed through sampling:  Rather than purchase each stock in the index, the ETF will instead develop a portfolio comprised of only a representative subset of the stocks in the index \cite{fuller2008evolution}.

The unsupervised stepwise selection algorithms provide a natural method for developing sampled portfolios:  To characterize the performance of an index of $n$ stocks over $m$ days, one can consider a data matrix $X$ containing the trace of each stock's daily percentage change in price\footnote{The daily percentage change in price is a summary statistic motivated by the logarithmic random walk model of stock price dynamics.  In this model, stocks are modeled to increase in price at an average multiplicative rate, but are subject to random fluctuations that are also multiplicative \cite{bachelier1900theorie, malkiel1999random}.},
\begin{eqnarray} \label{percentage_lift}
X_{ij} \equiv \frac{p_{i,j} - p_{i, j-1}}{p_{i, j-1}}.
\end{eqnarray}
Here, the row index $i$ identifies the stock and the column index $j$ the date.  Plugging (\ref{percentage_lift}) into the unsupervised selection algorithms, one obtains subset selections $s$ of the rows that approximately minimize the cost function (\ref{projection_error}).  The small $\vert s \vert$ subsets obtained in this way provide excellent candidates for inclusion in an index ETF, in that they will each contain stock selections that can be used to develop linear fits to each of the other stocks in $X$ with relatively high accuracy.   The greedy selection processes will naturally work towards selecting highly-representative subsets characterized my minimal redundancy,  resulting in well-diversified portfolios.

To illustrate the above strategy, I have applied the unsupervised selection algorithms to characterize the performance of $n=200$ of the highest market-cap health-care related stocks that traded on NASDAQ throughout the $m=252$ trading days of 2016\footnotemark.  The results of this analysis are summarized in Fig.\ \ref{fig:stocks}.  In the main plot of Fig.\ \ref{fig:stocks}a, I show the squared projection error that results when only the top $x$ stocks are retained, as output by the UFS model.  This shows that if one selects only the top 10, 25, 50, or  100 stocks ranked by this model,  41\%, 53\%, 68\%, and 83\% of the variance within the full 200 stock index is captured, respectively.  Elbows appear in this plot around the $10$ and $150$ stock selection marks.  This means that the improvement in variance captured is somewhat incremental after the first $10$ stocks, and is minimal in the last $50$.  The inset to  Fig.\ \ref{fig:stocks}a shows a plot of the improvement in $\mathcal{F}$ that results if we switch from the UFS algorithm to either the UFRS (dashed) or URS algorithms (solid) at different subset sizes.  As expected, the URS algorithm performs better at large $\vert s \vert$, and the hybrid UFRS algorithm performs better throughout.

\footnotetext{The 200 health-sector stock tickers included in the analysis are: JNJ, PFE, NVS, MRK, UNH, AMGN, MMM, SNY, MDT, GSK, ABBV, CELG, NVO, WBA, LLY, BMY, GILD, CVS, AGN, AZN, ABT, BIIB, SYK, ANTM, AET, ESRX, CI, BDX, REGN, BSX, TEVA, HCA, HUM, ISRG, VRTX, MCK, BAX, ZTS, LUX, ALXN, FMS, INCY, ZBH, CAH, EW, MYL, Q, BCR, ABC, BMRN, LH, SNN, XRAY, DGX, SHPG, IDXX, HSIC, DVA, ANTX, CNC, HOLX, UHS, RMD, COO, PRGO, ALGN, SGEN, JAZZ, TFX, ALKS, VAR, TSRO, RDY, WCG, QGEN, EVHC, DXCM, EXEL, STE, MD, HLF, IONS, ABMD, UTHR, MASI, TARO, HRC, MNK, NBIX, KITE, ICLR, ALNY, ALR, CRL, PDCO, OPK, AKRX, RAD, PRAH, ACAD, HLS, NUVA, TECH, ACHC, MDCO, CTLT, BLUE, HCSG, PRXL, WMGI, CBPO, CMD, CHE, GMED, PBH, ICUI, NUS, VRX, GRFS, PEN, NKTR, MOH, SAGE, ICPT, EXAS, GWPH, MSA, RARE, JUNO, NVRO, LIVN, HZNP, LPNT, CLVS, BKD, PODD, XON, INCR, IRWD, NEOG, AXON, ENDP, ZLTQ, LGND, PTLA, HAE, OMI, TBPH, PRTA, AGIO, CBM, NXTM, SEM, PCRX, SRPT, AMED, HYH, HALO, FGEN, MGLN, RDUS, ONCE, INGN, SUPN, GKOS, LXRX, THC, BPMC, AAAP, MMSI, TDOC, ARRY, AERI, PBYI, INVA, EGRX, DERM, BABY, CNMD, MDXG, GBT, LOXO, MYGN, SPNC, RGEN, EBS, FMI, INSM, TVTY, NHC, PAHC, XNCR, XLRN, GHDX, DPLO, FOLD, ALDR, CORT, NRCIB, NVCR.  Historical closing values were obtained from http://real-chart.finance.yahoo.com.}

The results in Fig.\ \ref{fig:stocks}b summarize the results that are obtained if one applies the algorithms to a five-day moving average of (\ref{percentage_lift}).  This smoothing serves to reduce short time-scale noise that is often fairly idiosyncratic.  In this case, retaining the top 10, 25, 50, or 100 stocks from the UFS procedure results in capturing 47\%, 66\%, 84\%, and 96\% of the full index variance, respectively.  Averaging over longer periods results in  further noise reduction, allowing for even better performance at small $\vert s \vert$.  For example, averaging over a 30 day window, the top 10, 25, 50, or 100 stocks selected by the forward procedure capture 80\%, 92\%,  97\%, 99\% of the index variance, respectively.  An elbow in the plot (not shown) now appears around the 25 stock mark, motivating a portfolio selection of this size.

The example of sampled, representative portfolio development illustrates well the benefits one receives through application of the unsupervised selection procedures:  Primarily, these procedures allow for an automated dimensional reduction to be effected, reducing the number of features needed to accurately characterize the behavior of a system.  Automated feature-engineering-based dimensional reduction methods are also available for this purpose, but these are not always appropriate.  For example, in the case of ETF development, we explicitly require the selection of a representative subset of the stocks -- the identification of a set of informative, hybridized stocks that include each of those present would provide little utility.  The analysis considered here has also demonstrated the potential for improved fits to be obtained through use of the URS and UFRS algorithms in different limits.

\section{Discussion}
\label{sec:discussion}

In this paper, I have presented unsupervised extensions of the fast stepwise linear regression algorithm.  These extensions allow one to quickly identify  feature subsets that are highly representative of the whole.  In colloquial terms, one can think of these algorithms as feature selection analogs of the popular principal component analysis (PCA) algorithm.  Whereas PCA identifies the unrestricted linear combinations of features that capture the most variance within a data set, the algorithms presented here identify subsets of the original features that do so.  This approach retains the original interpretability of the resulting features within the dimensionally-reduced data set, which can be important for certain applications.  Like PCA, the greedy stepwise algorithms center around consideration of the feature set correlation matrix.  The number of computations required to evaluate this correlation matrix sets the shared runtime for each, which is quadratic in the number of features and linear in the number of data samples available.  This efficiency makes the fast stepwise algorithms compelling candidates for feature selection tasks at scale.

I discuss two interesting properties relating to these algorithms in the appendices.  In Appendix \ref{sec:error}, I have shown that the error in the matrices determining which feature should next be pruned or added to a predictor set grows slowly with iteration -- this error scales like the square root of the iteration index.  This slow growth of error allows one to work with relatively low numerical precision in many applications, which can result in both reduced run-times and reduced memory requirements.  Second, in Appendix \ref{sec:relationship}, I have shown that the feature orderings returned  by the forward and reverse algorithms are exactly opposite whenever the features present are all the same, up to some fluctuations that are only weakly correlated.  The result is somewhat surprising, but represents the correct greedy behavior for the optimization of (\ref{projection_error}) in the highly-redundant limit.  When the features present are not weakly correlated, or are not all highly-redundant, the relationship between the two protocols is certainly more complex.

An alternative formalism for carrying out unsupervised forward stepwise regression was introduced in \cite{farahat2013efficient}.  This prior work also introduced a distributed implementation of the forward algorithm that allows for applications to very large data sets.  It would be useful to develop a distributed version of the formalism considered here, or alternatively to extend the formalism of  \cite{farahat2013efficient} to support the reverse and hybrid forward-reverse protocols.  One benefit favoring the formalism considered here is that it allows for the possibility of quick development of high quality implementations via simple extensions of already existing supervised implementations.

\appendices

\section{Correlation matrix inverse updates}
\label{sec:derivation}
Efficient stepwise regression centers around fast updating of the correlation matrix inverse as the predictor set considered is iteratively altered:  Typically, an $n \times n$ matrix requires $O(n^3)$ operations to invert, but using the update approach, the correlation matrix inverse can be corrected at each step of these algorithms in only $O(n^2)$ time \cite{efroymson1960multiple}.  For completeness, I present a derivation of the update formulas needed to do this here.  This is based upon application of the Woodbury inverse update formula \cite{woodbury1950inverting}:  Given an initial $n \times n$ matrix $M$ and a second $n \times n$ matrix  $u  v^T$ -- with $u$ and $v$ both $n\times k$ matrices -- the formula states
\begin{eqnarray}\label{Woodbury} 
\lefteqn{(M + u  v^T)^{-1} =} \\&& M^{-1} - M^{-1}  u  \left (I + v^T  M^{-1}  u \right )^{-1}  v^T M^{-1}. \nonumber
\end{eqnarray}
The correctness of this expression can be confirmed through direct multiplication.  

To apply (\ref{Woodbury}) here, I take $M$ to be our correlation matrix,
\begin{eqnarray}
 M = \begin{bmatrix}
    M_{11}  & \dots &  M_{1n} \\
    \hdotsfor{3} \\
    M_{n1}  & \dots  & M_{nn}
\end{bmatrix},
\end{eqnarray}
with $M_{ij}$ given by (\ref{correlation}).  Writing
\begin{eqnarray} \label{uv1}
u = - \begin{bmatrix}
    0      & M_{1n} \\
    0      & M_{2n} \\
    \hdotsfor{2} \\
    0      & M_{(n-1)n} \\
    1     &  0,      
\end{bmatrix},  \ \ \
v = 
\begin{bmatrix}
    M_{1n}  & 0 \\
    M_{2n} &0 \\
    \hdotsfor{2} \\
    M_{(n-1)n} &0 \\
    0     &  1   
\end{bmatrix},
\end{eqnarray}
one obtains
\begin{eqnarray} \label{Miter}
M + u v^T =
\begin{bmatrix}
    M_{11}        & \dots & M_{1(n-1)} & 0 \\
    M_{21}        & \dots & M_{2(n-1)} & 0\\
    \hdotsfor{4} \\
    M_{(n-1)1}   & \dots & M_{(n-1)(n-1)} & 0\\
    0     &   \dots & 0 & 1       
\end{bmatrix}.   
\end{eqnarray}
The upper-left sub-matrix here is the correlation matrix of the first $(n-1)$ features.  Its inverse can be obtained from the right side of (\ref{Woodbury}).  To evaluate this expression, first note that the definition of $M^{-1}$ gives
\begin{eqnarray}
M^{-1}  u =   \begin{bmatrix}
    -M^{-1}_{1n}      & M^{-1}_{1n} \\
    -M^{-1}_{2n}     & M^{-1}_{2n} \\
    \hdotsfor{2} \\
    -M^{-1}_{(n-1)n}      & M^{-1}_{(n-1)n} \\
   - M^{-1}_{nn}     &  M^{-1}_{nn} -1     
\end{bmatrix}, \label{left_matrix}
\end{eqnarray}
and
\begin{eqnarray}
v^T M^{-1}  = \begin{bmatrix}
    -M^{-1}_{1n} &  \cdots &- M^{-1}_{(n-1)n} & -M^{-1}_{nn} + 1 \\
    M^{-1}_{1n}   & \cdots & M^{-1}_{(n-1)n} & M^{-1}_{nn}
\end{bmatrix}. \label{right_matrix}
\end{eqnarray}
Similarly, we have 
\begin{eqnarray}
\left (I + v^T M^{-1}  u \right)^{-1} = \begin{bmatrix}
   1       & 1 - \frac{1}{(M^{-1})_{nn}}\\
   1     &  1
\end{bmatrix}. \label{inner_matrix}
\end{eqnarray} 
Using these last three lines, and plugging into (\ref{Woodbury}) gives
\begin{eqnarray}\label{reverse_inverse_update}
(M+ u  v^T)^{-1}  =  M^{-1}  - \frac{M^{-1}_n \otimes M^{-1}_n }{M^{-1}_{nn}} + e_{nn},
\end{eqnarray}
where $M^{-1}_n$ is the $n$-th row of $M^{-1}$, $\otimes$ represents the outer product, and $e_{nn}$ is the $n \times n$ matrix has a $1$ in the $n$-th diagonal and is zero elsewhere.  The components outside of $n$ give the inverse of the correlation matrix of the first $n-1$ features.  Note that (\ref{reverse_inverse_update}) can be evaluated in $O(n^2)$ time, given knowledge of $M^{-1}$, the correlation matrix of the first $n$ components.

One can obtain the inverse update formula that applies when a feature is added to the retained set by reversing the above steps.  This is done by changing the sign of $u$ in (\ref{uv1}).  Carrying out similar algebra to that above gives the update for adding feature $n$ back to $s$,
\begin{eqnarray}\label{forward_inv_update} 
\lefteqn{M^{-1} \leftarrow } \\ && M^{-1} + \frac{\left ( M^{-1} \cdot M_n - e_n \right) \otimes \left ( M^{-1} \cdot M_n - e_n \right) }{1 - M_n^T \cdot M^{-1} \cdot M_n}. \nonumber
\end{eqnarray}
Here, $e_n$ is the vector that is zero everywhere except index $n$ where it takes value $1$, $M^{-1}$ is now the inverse correlation matrix of the first $n-1$ features, and $M$ is the full correlation matrix.  These results are equivalent to (\ref{updateformula}) and (\ref{forward_inverse_update}).

\section{Error propagation}
\label{sec:error}

A potential concern associated with applying iterative algorithms is the possibility of compounding growth of numerical error with each iteration.  However, I show here that error grows slowly with iteration count in the unsupervised selection algorithms.  To illustrate, consider the situation where the reverse process is initialized with a numerical approximation to the true inverse of $M$ given by
\begin{eqnarray}\label{error1}
\mu^{-1}  = M^{-1} + \epsilon,
\end{eqnarray}
where $\epsilon$ is a small error term.  The error at later steps in the process will contain contributions from this initial error, as well as contributions from error introduced through later numerical roundoff mistakes.  Consider first the idealized error that is a consequence of the initial error in (\ref{error1}) only.  The correlation matrix associated with (\ref{error1}) is given by its inverse.    To leading order in $\epsilon$, this is
\begin{eqnarray}\nonumber \label{error2}
\mu  &= & \left ( M^{-1} + \epsilon \right )^{-1}  \\
&=&M- M \epsilon M +O(\epsilon^2),
\end{eqnarray}
a result following from the Taylor expansion of the inverse \cite{apostol1969calculus}.  At a later step in the reverse process -- again, ignoring round-off error for the moment -- the inverse that obtained will be equal to the exact inverse of a feature-pruned version of (\ref{error2}).  That is, a matrix $\mu^{\prime}$ obtained from (\ref{error2}) by simply setting some its rows and columns to zero.  The exact inverse of this matrix will be given by
\begin{eqnarray}\nonumber \label{idealizederror}
\mu^{\prime -1} &=& \left (M^{\prime}- \left(M \epsilon M \right)^{\prime} +O(\epsilon^2)\right )^{-1} \\ 
&=& M^{\prime -1} + M^{\prime -1} \left(M \epsilon M \right)^{\prime}  M^{\prime -1}  + O(\epsilon^2).
\end{eqnarray}
This result shows that the leading error in the output of the idealized algorithm -- i.e.\ that not subject to continued introduction of roundoff error -- will remain small after any number of iterations.  In other words, there is no build-up, or concentration of the initial error in the final components of the pruned correlation matrix inverse.

\begin{figure}[t]\scalebox{0.60}
{\includegraphics[angle=0]{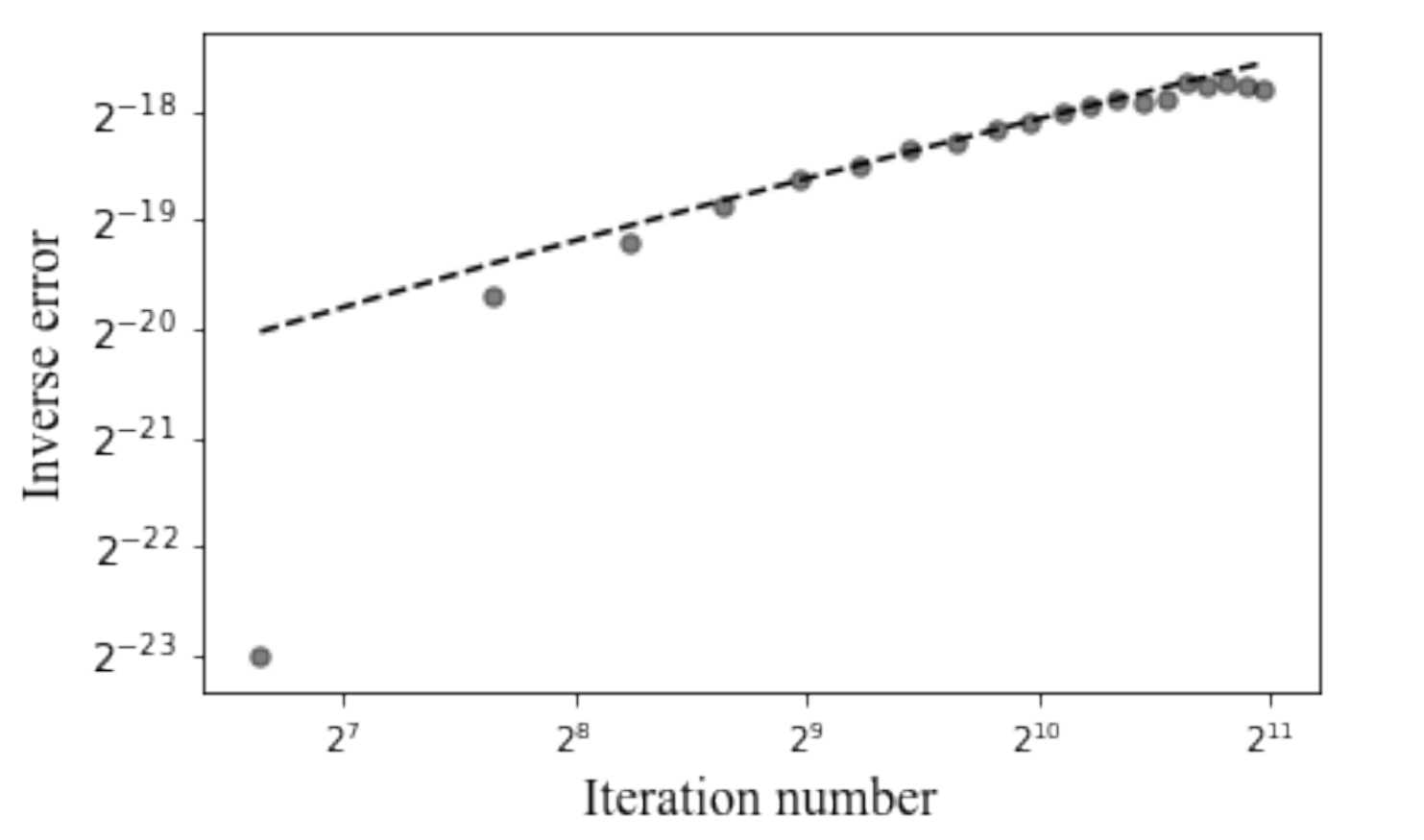}}
\caption{\label{fig:error} Log-log plot of element-wise maximum absolute error in $M^{-1}$ versus reverse algorithm iteration.  The data points represent averages over ten runs of the algorithm, each time beginning with a data set consisting of $n= 2000$ random features, each with $m=8000$ data points.  The error was evaluated through comparison of the algorithm's output for the inverse after $k$ steps  and that obtained through direct evaluation of the inverse with the corresponding pruned features removed.  The dashed line represents a fit to the data using form (\ref{error3}) -- that is, it has slope $1/2$.  The quality of the fit suggests consistency with the random-walk model outlined in the text.}
\end{figure}

To relate (\ref{idealizederror}) to the error of a realistic implementation, one needs to consider the numerical roundoff introduced with each iteration.  To leading order, these errors are additive.  Further, one can expect these errors to be relatively uncorrelated across iterations,  implying that the net error growth after $k$ steps should scale like that for a random walk \cite{rudnick1987shapes},
\begin{eqnarray}\label{error3}
\text{Error after $k$ steps } \sim O(\epsilon \times \sqrt{k}).
\end{eqnarray}
Here, I now use $\epsilon$ for the scale of the round off error.  Numerical experiments confirm this expected scaling -- see Fig.\ \ref{fig:error}.  Similar square-root scaling also characterizes the error growth in the iterative $D$ and $U$ updates returned by the algorithm.

A consequence of the small error growth rate (\ref{error3}) is that accurate feature orderings can be obtained from the algorithm without requiring high precision variable types.  This can be very helpful when working with systems having a great many features, as working with smaller precision floats can significantly reduce run time and memory requirements.  In general, one can use (\ref{error3}) to select the precision level needed for a given procedure.  For example, if $N \sim 10^4$ and the maximum desired error is $E = O(10^{-5})$, then floating point arithmetic valid at $\epsilon \sim O(E / \sqrt{N}) \sim O(10^{-7})$ will suffice.  A trade-off between precision and speed can also be made:  Smaller precision variables can be used if one is willing to occasionally reset the error in the inverse through direct -- i.e.\ $O(n^3)$ complexity -- re-evaluation.

\section{Relation between forward and reverse solutions}
\label{sec:relationship}

In the supervised context -- where one attempts to find the optimal feature subset of size $k$ for a linear fit to a target variable $y$ --  both the forward and reverse greedy algorithms have been analyzed in detail.  The reverse algorithm has been shown to always return the optimal feature subset, provided the target variable  $y$ is not too far from the space spanned by the optimal $k$ regressors \cite{harikumar1998fast, couvreur2000optimality}.  The same conclusion does not hold for the forward algorithm.  However, it has been shown that if an exact, sufficiently sparse representation of the target variable exists within the space spanned by the available features, the forward algorithm will find it   \cite{tropp2004greed, bruckstein2009sparse}.  Both of these results are quite significant, given that the challenge of identifying the global optimum is in general an NP-hard task \cite{natarajan1995sparse}.  These results suggest that both the forward and reverse greedy approaches might also be expected to perform well in the unsupervised learning context, provided the features in question are relatively redundant -- so that each can be well fit by some subset of the others.

An interesting limiting case is provided by the situation where all feature fluctuations are only weakly correlated.  In this regime, one can show that the forward and reverse algorithms return exactly opposite feature orderings:  Consider a correlation matrix of the form
\begin{eqnarray}
M = I + \epsilon N,
\end{eqnarray}
where $\epsilon$ is a small number and $N$ is a symmetric, fully off-diagonal matrix with elements of size $O(1)$.  If some subset $s$ of the features is selected, the regression error on the remaining features can be read off from (\ref{rjsquared}).  Expanding, this is
\begin{eqnarray}\nonumber \label{forward_backward_1}
\mathcal{F}_s -  \left ( n - \vert s \vert \right)  &=& - \sum_{j \not \in s}  \epsilon N_j^T \cdot M^{-1} \cdot \epsilon N_j \\
&=&  - \epsilon^2 \sum_{i \in s} \sum_{j \not \in s} N_{ij}^2 + O(\epsilon^3).
\end{eqnarray}
In the first line here, $N$ is  again the full matrix and $N^{-1}$ is the updated inverse, which is non-zero only over the columns within $s$. 

The result (\ref{forward_backward_1}) shows that to leading order the projection error is the sum of the squared elements of the block off-diagonal elements of $M$ -- those elements where one index is in $s$ and one is not.  By symmetry -- at the same order of approximation -- this must also be the squared projection error observed when $s$ is projected onto its complement.  If the forward process is run from this starting point -- from the perspective of $s$ -- the next feature selected will be that that returns the largest sum over the resulting block off-diagonal elements squared.  But by the symmetry noted above, if the reverse process is run from this same starting point -- from the perspective of the complement of $s$ -- the same feature would also be selected for the next removal from the complement, as this element will also minimize the squared error of projecting $s$ onto its complement.  In this way, the two algorithms will result in completely opposite orderings:  The first feature selected by the forward process will also be the first pruned by the reverse process, etc.  It's important to stress that these opposite orderings are not a consequence of errors in the algorithms in question, but instead represent the correct greedy responses to weakly correlated features.  A similar result will also occur if all features take the form $X_i = \overline{X} + \delta X_i$, where $\overline{X}$ is a vector common to all features, but the $\delta X_i$ vectors are only weakly correlated.  This is a model that may be appropriate for many feature clusters.

\section*{Acknowledgement}
The author thanks  J.\ Bergknoff,  E.\ Jeske, P.\ Spanoudes, D.\ Thorpe, C.\ Yeh, and R.\ Zhou for helpful discussions relating to this work.

\bibliographystyle{abbrv}  
\bibliography{refs}

\begin{thebibliography}{10}

\bibitem{apostol1969calculus}
T.~M. Apostol.
\newblock Calculus: Multivariable calculus and linear algebra, with
  applications to differential equations and probability, 1969.

\bibitem{bachelier1900theorie}
L.~Bachelier.
\newblock {\em Th{\'e}orie de la sp{\'e}culation}.
\newblock Gauthier-Villars, 1900.

\bibitem{bruckstein2009sparse}
A.~M. Bruckstein, D.~L. Donoho, and M.~Elad.
\newblock From sparse solutions of systems of equations to sparse modeling of
  signals and images.
\newblock {\em SIAM review}, 51(1):34--81, 2009.

\bibitem{couvreur2000optimality}
C.~Couvreur and Y.~Bresler.
\newblock On the optimality of the backward greedy algorithm for the subset
  selection problem.
\newblock {\em SIAM Journal on Matrix Analysis and Applications},
  21(3):797--808, 2000.

\bibitem{efroymson1960multiple}
M.~Efroymson.
\newblock Multiple regression analysis.
\newblock {\em Mathematical methods for digital computers}, 1:191--203, 1960.

\bibitem{farahat2013efficient}
A.~K. Farahat, A.~Ghodsi, and M.~S. Kamel.
\newblock Efficient greedy feature selection for unsupervised learning.
\newblock {\em Knowledge and information systems}, 35(2):285--310, 2013.

\bibitem{fuller2008evolution}
S.~L. Fuller.
\newblock The evolution of actively managed exchange-traded funds.
\newblock {\em The Review of Securities and Commodities Regulation},
  41(8):89--96, 2008.

\bibitem{guyon2003introduction}
I.~Guyon and A.~Elisseeff.
\newblock An introduction to variable and feature selection.
\newblock {\em Journal of machine learning research}, 3(Mar):1157--1182, 2003.

\bibitem{harikumar1998fast}
G.~Harikumar, C.~Couvreur, and Y.~Bresler.
\newblock Fast optimal and suboptimal algorithms for sparse solutions to linear
  inverse problems.
\newblock In {\em Acoustics, Speech and Signal Processing, 1998. Proceedings of
  the 1998 IEEE International Conference on}, volume~3, pages 1877--1880. IEEE,
  1998.

\bibitem{malkiel1999random}
B.~G. Malkiel.
\newblock {\em A random walk down Wall Street: including a life-cycle guide to
  personal investing}.
\newblock WW Norton \& Company, 1999.

\bibitem{masaeli2010convex}
M.~Masaeli, Y.~Yan, Y.~Cui, G.~Fung, and J.~G. Dy.
\newblock Convex principal feature selection.
\newblock In {\em Proceedings of the 2010 SIAM International Conference on Data
  Mining}, pages 619--628. SIAM, 2010.

\bibitem{natarajan1995sparse}
B.~K. Natarajan.
\newblock Sparse approximate solutions to linear systems.
\newblock {\em SIAM journal on computing}, 24(2):227--234, 1995.

\bibitem{rudnick1987shapes}
J.~Rudnick and G.~Gaspari.
\newblock The shapes of random walks.
\newblock {\em Science}, 237:384--390, 1987.

\bibitem{tropp2004greed}
J.~A. Tropp.
\newblock Greed is good: Algorithmic results for sparse approximation.
\newblock {\em IEEE Transactions on Information theory}, 50(10):2231--2242,
  2004.

\bibitem{woodbury1950inverting}
M.~A. Woodbury.
\newblock Inverting modified matrices.
\newblock {\em Memorandum report}, 42:106, 1950.

\end{thebibliography}

\end{document}